\documentclass{article}
\usepackage[utf8]{inputenc}
\usepackage{spconf,amsmath,graphicx}
\usepackage{float}
\usepackage{subfig}
\usepackage{multirow}

\title{Self-supervised learning for segmentation} %kidney 
\name{Abhinav Dhere, Jayanthi Sivaswamy}
\address{Center for Visual Information Technology (CVIT), IIIT-Hyderabad, India}
\begin{document}

\maketitle
\begin{abstract}
Self-supervised learning is emerging as an effective substitute for transfer learning from large datasets. In this work, we use kidney segmentation to explore this idea. The anatomical asymmetry of kidneys is leveraged to define an effective proxy task for kidney segmentation via self-supervised learning. A siamese convolutional neural network (CNN) is used to classify a given pair of kidney sections from CT volumes as being kidneys of the same or different sides. This knowledge is then transferred for the segmentation of kidneys using another deep CNN using one branch of the siamese CNN as the encoder for the segmentation network. Evaluation results on a publicly available dataset containing computed tomography (CT) scans of the abdominal region shows that a boost in performance and fast convergence can be had relative to a network trained conventionally from scratch. This is notable given that no additional data/expensive annotations or augmentation were used in training.   
\end{abstract}


\section{Introduction}

In nature, human beings learn cumulatively. We build upon previous knowledge to learn new concepts quickly and efficiently. 
% This idea is mimicked 
One way to achieve this in neural networks has been through transfer learning where a pre-trained (on a large dataset) network is used for training for the new task with the associated data. Generally, a neural network learns basic features such as edges and lines in the first few layers. The complexity of features learned by the network increases from the input layer to the final layer with the later ones being more task-specific. Hence, only the weights of the later layers are typically modified during transfer learning.
Ideally, the similarity level of the pre-training task and main task can yield a good transfer in performance. However, in the case of medical images where labeled data is scarce, this is difficult. Nevertheless, there have been attempts to do transfer learning across domains, for example, disease classification with the VGG network \cite{menegola2017knowledge} whereas such attempts are sparse for the segmentation task.  

% ImageNet is a large dataset containing millions of images with manual annotations for classification and object recognition ~\cite{deng2009imagenet}. In computer vision, it is a common practice to use networks pre-trained on ImageNet for transfer learning. However, these images are collected from the internet and annotated by crowdsourcing. This makes the dataset very different from medical images, in general. Intuitively, it seems reasonable that for a task in medical image analysis, only the basic features learned from such a dataset could be transferred.  

Self-supervised learning is another recent paradigm that can offer a solution to the data sparsity problem. Here, a model is pre-trained on a proxy task based on a dataset that is also used for the main task. This proxy task is chosen such that it is easy to solve, but requires learning some form of image semantics. It is formulated such that implicit information in the same dataset can be used as labels, thus removing the need for additional annotation. 
 
Many examples of such tasks are found in the recent literature in computer vision. Agrawal et al.~\cite{agrawal2015learning} tasked the network to predict the camera transformation of moving targets for the main tasks of scene recognition and recognition. Doersch et al.~\cite{doersch2017multi} predicted the relative position of patches for object detection in the Pascal VOC dataset \cite{everingham2010pascal}. Recently, Vondrick et al.~\cite{vondrick2018tracking} used the colorization of videos as a proxy task for visual tracking. 
% solve a jigsaw puzzle \cite{noroozi2016unsupervised} etc. 

Self-supervision is a promising idea in medical image analysis since medical data is often rich in anonymous metadata such as patient age, medical history, and longitudinal information but often lacking manual annotations by medical experts.
On longitudinal MR data, Jamaludin et al.~\cite{jamaludin2017self}  predicted the degradation level of vertebral disks using self-supervision after using a proxy task which classified pairs of images as belonging to the same or a different patient. Li et al.~\cite{li2018non} maximized the distance between registered images at multiple scales as a proxy task for non-rigid image registration.
However, the use of self-supervision for improving segmentation in medical volumes is relatively unexplored.  
 
In this work, we explore self-supervision for medical image segmentation. The selection of an appropriate proxy task is critical to the effectiveness of self-supervised learning and needs careful design. We demonstrate this for kidney segmentation from abdominal CT volumes. The proposed proxy task exploits the anatomical \textit{asymmetry} between the left and right kidneys. We show that segmentation with self-supervision allows for faster convergence and a better performance relative to training from scratch while using the same amount of data. 

\section{Method}
\begin{figure}
    \centering
    \includegraphics[scale=0.35]{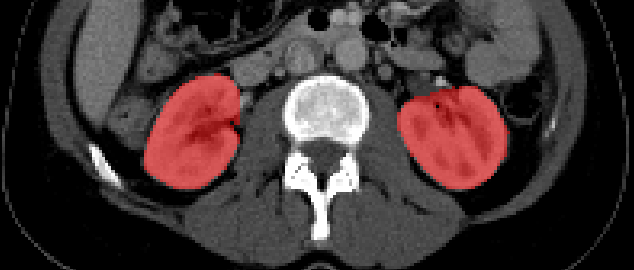}
    \caption{Axial view of right and the left kidney in a CT volume}
    \label{fig:kidney}
\end{figure}

\begin{figure}
     \centering
     \includegraphics[scale=0.2]{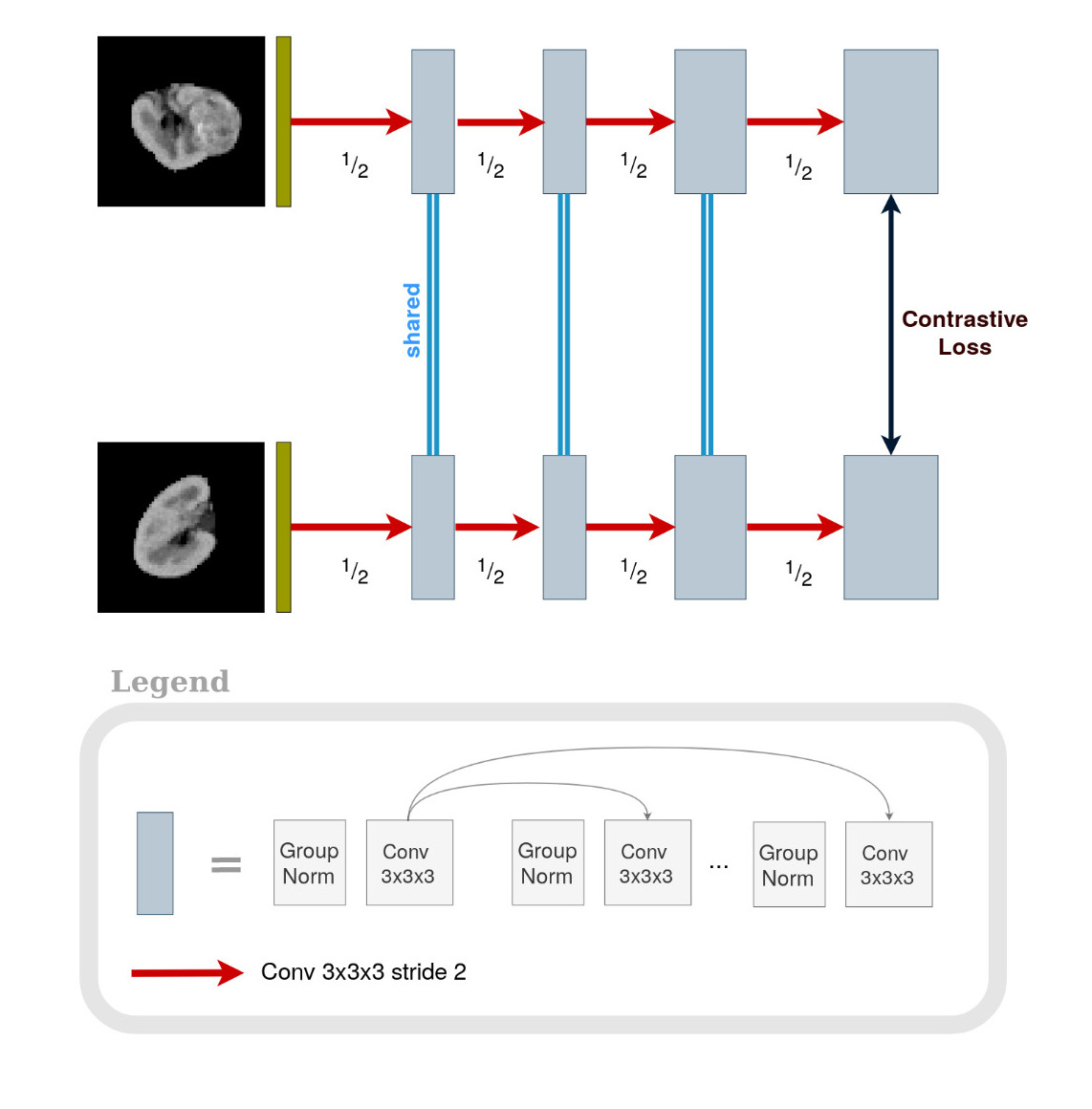}
     \caption{Siamese network for proxy task}
     \label{fig:proxyArch}
 \end{figure}

The left and right kidney differ slightly in size, shape, and spatial location~\cite{frimann1961normal} (see fig \ref{fig:kidney}). Thus, we propose a proxy task to classify whether a given pair of kidneys belong to the same side. To accomplish this, the network needs to develop an understanding of the structure and sizes of the kidneys. We hypothesize that this knowledge is useful in the main task of kidney segmentation. 
We begin by dividing the abdominal CT volume into two halves along the sagittal plane, thus obtaining one kidney in each half. We observed that when complete halves of the CT scan are passed, the network learns to perform this task very easily, based on the spatial location of the two kidneys. In such cases when the task is too easy and does not require considerable semantics, it is observed that transfer learning to do a main task is poor. It is desirable to force the network to learn to do the task, based only on the shape and size, rather than the spatial location.
Accordingly, each kidney section is extracted using the kidney annotation mask and a fixed region of 64x112x112 pixels around the kidney is passed to the network.
% Accordingly, a bounding box is extracted around the kidney and the kidney volume (in each half) is identified via an intensity-based thresholding step. This can be achieved using the kidney annotation mask. \textbf{if annotation is available why do you need to do all the above steps?}
Thus, only the kidney part of the scan is passed to the network for the proxy task.
 A siamese network, composed of two identical sub-networks with shared weights~\cite{bromley1994signature}, is employed to perform this proxy task. Each of the two branches of the network is given the region of one kidney. Label 1 (0) is assigned if both the kidneys are of the same (opposite) side. The siamese network for the proposed proxy task is shown in fig. \ref{fig:proxyArch}
The network is trained using contrastive loss, defined as 
\begin{equation*} 
L_C=\sum^N_{n=1} (y)d^2+ (1-y) \text{max}(0,m-d)^2 
\end{equation*} 
where, $y$ is the label; $d = || a - b ||^2 $ is the L2 distance between the embeddings $a$ and $b$ of the final layers of each branch of the siamese network; $m$ is the margin.
%up to which the distance should be maximized when opposite pairs for the embeddings to be considered as perfectly trained. 
This loss works by minimizing the distance when the label is 1 but maximizing it when the label is 0. After the embeddings of opposite pairs cross $m$, the loss flips to 0. The value of margin was set to 1 in our implementation.

On completion of training for the proxy task, one of the branches of the siamese network is used as an encoder for the main kidney segmentation task. Since the siamese network uses shared weights for both branches, any of the branches can be used as the encoder. Thus, only the encoder part of the segmentation network is pre-trained. The rest of the network uses randomly initialized weights. The segmentation network is trained using a weighted sum of soft dice loss~\cite{milletari2016v} and weighted binary cross-entropy (BCE). The weights for BCE and dice loss are initialized as 0.6 and 0.4 respectively. As training progresses, the weight for BCE is decreased while the weight for dice loss is increased by the same amount. The intuition behind this is that BCE improves voxelwise performance while dice loss improves segmentation overlap. 

\section{Experiments and results}
 
\subsection{Implementation details}
We employ an encoder-decoder system with U-Net like skip connections for the segmentation. The encoder branch consists of 4  blocks with densely connected layers. It has been shown that densely connected layers control the vanishing gradient problem while keeping the number of parameters low~\cite{huang2017densely}. Each of these blocks consists of 2, 2, 4 and 8 densely connected layers respectively. After each block, downsampling is performed with the help of a 3D convolutional layer with stride 2. This is followed by another convolutional layer with kernel size 1x1x1 to combine and reduce the number of feature maps.

\begin{figure}[h!]
    \centering
    \includegraphics[scale=0.35,trim={1.2cm 0 0 0}]{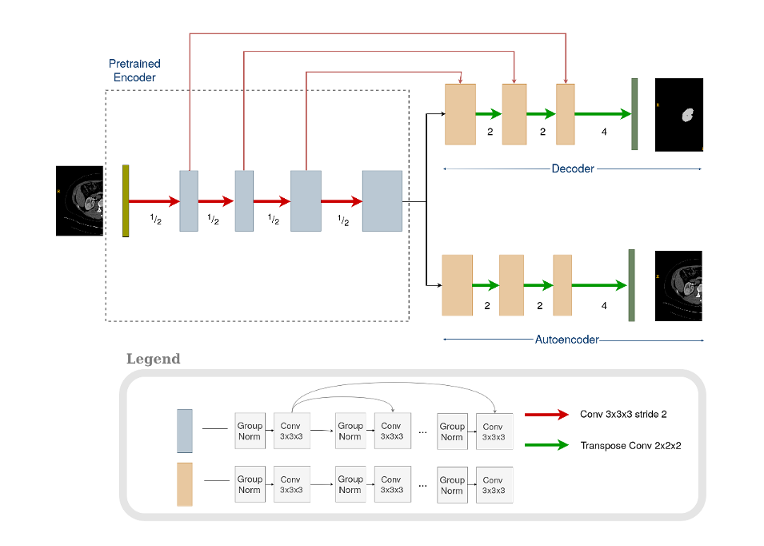}
    \caption{Network architecture for segmentation}
    \label{fig:DUN}
\end{figure}

To reduce complexity, the decoder is smaller and consists of plain 3D convolutional blocks connected serially. Each block in the decoder comprises of 2 3D convolutional layers of kernel size 3x3x3. After each block, upsampling is performed using a transpose convolutional layer. 
We also connect an additional reconstruction section to the encoder to form an autoencoder. This reconstruction part is structurally similar to the decoder except for the last layer but does not employ any skip connections. This block serves as a regularizer during training and controls overfitting~\cite{sabour2017dynamic}. Fig.~\ref{fig:DUN} shows the network architecture.

As 3D convolution is memory intensive, we have trained with batch size 1. Group norm is better suited for regularization in the case of small batch sizes~\cite{groupNorm}. Hence, all convolutional layers are preceded by a Group norm layer. We use Adam optimizer with an initial learning rate $10^{-3}$ and weight decay $10^{-2}$. The code is implemented with the PyTorch framework in Python and was run on NVIDIA GTX 1080GPU with 11GB RAM.

\subsection{Dataset}
The CT dataset from the KiTS 2019 challenge~\cite{kitsData} was used for assessment. It consists of abdominal CT scans of 210 cases with manual annotations and 90 cases without annotation. The data was first resampled to have a uniform voxel spacing of 3.22x1.62x1.62 mm, resulting in a median size of 128x248x248 voxels. The voxels were clipped to a window of [-80,300] by a windowing operation to improve the contrast of the kidney region. Data normalization was done by subtracting the mean and dividing by the standard deviation. Rather than resizing the volumes to a fixed size, the volumes were zero-padded to obtain a size closest to the nearest multiple of 16 to permit some variable size input to the network.

\subsection{Experiments and results}
\begin{figure}[h!]
    \includegraphics[scale=0.4]{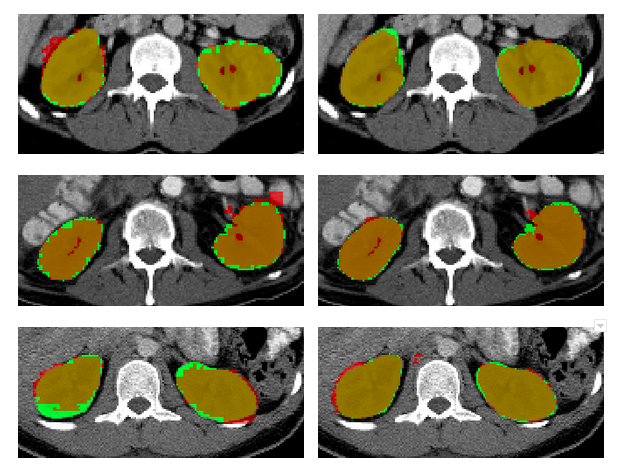}
    \caption{\label{fig:segImage} Sample segmentation results at the end of 200 epochs for 3 cases. Top to bottom: cases 1-3. Left to Right: results of models trained without (i.e. trained from scratch) and with self-supervision, overlaid on the ground truth. Green: Ground truth; Red: Prediction.}
    % the ground truth, results of models trained without (i.e. trained from scratch) and with self-supervision.}
\end{figure}
The 210 volumes were split randomly into 2 sets: 80\% for training while reserving the remaining 20\% for validation. Two identical networks were taken and one was trained using self-supervision and the resulting model is henceforth referred to as a model with self-supervision or MwS. The second network was trained from scratch and results in a model referred to as a model without self-supervision or MwoS. The initial training settings for both networks were identical and training was done for a fixed number of epochs. The effectiveness of self-supervised learning was evaluated both qualitatively and quantitatively. 

\begin{figure}
    \centering
    \includegraphics[scale=0.3]{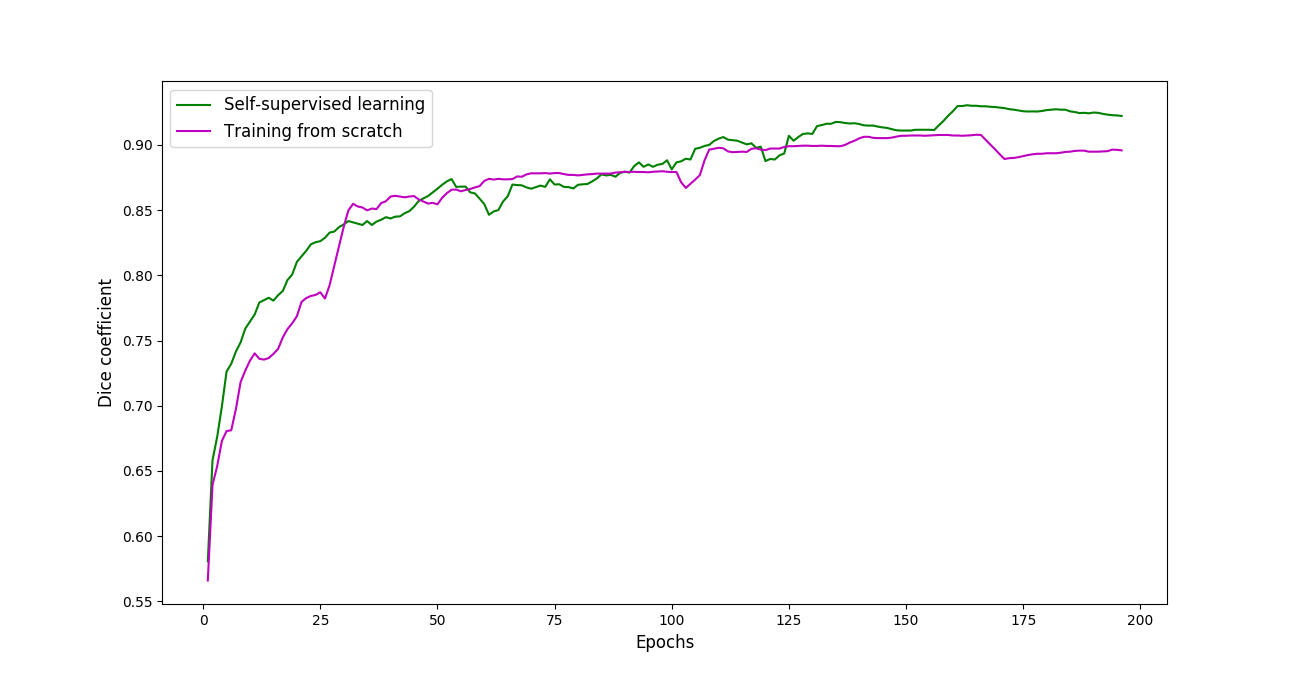}
    \caption{Progression of DC over epochs on the validation set}
    \label{fig:diceCurve}
\end{figure}

For the quantitative comparison, the mean Dice score coefficient (DC) was chosen. This score is defined as
\begin{equation*}
    DC = \frac{2TP}{2TP+FP+FN}
\end{equation*}
where TP, FP and FN are the number of true positives, false positives and false negatives respectively. 

\begin{table}
\centering
\caption{\label{tab:res}Comparison of quantitative results}

\begin{tabular}{|c|c|c|c|c|c|c|}
\cline{1-7} 
\multicolumn{1}{|c|}{\multirow{2}{*}{}} & \multicolumn{2}{c|}{DC} & \multicolumn{2}{c|}{HD} & \multicolumn{2}{c|}{BL (\%)}\tabularnewline
\cline{2-7} 
 & MwoS & MwS & MwoS & MwS & MwoS & MwS\tabularnewline
\hline 
\hline 
% Train. & 0.906 & 0.945 & 0.378 & 0.323 & 21.0 & 22.1\tabularnewline
Val. & 0.907 & 0.931 & 0.303 & 0.291 & 14.0 & 12.9\tabularnewline
Test & 0.853 & 0.880 & - & - & - & -\tabularnewline
\hline 
\end{tabular}

\end{table}
For further validation, we also compared two additional metrics -  Hausdorff distance (HD) and the difference in boundary length of prediction and ground truth (BL).
The obtained values (for both MwoS and MwS) for these are presented for the validation, and test sets in table \ref{tab:res}. 
After training for 200 epochs, a peak DC of 0.931 and 0.907 were achieved with MwS and MwoS, respectively, on the validation set.
Ideally, higher values (close to 1) are good for DC and lower values are good for HD and BL. 
It is evident from table \ref{tab:res} that MwS performs better than MwoS as per all 3 metrics.
The test set of 90 volumes is released by challenge organizers without annotations. Evaluation results (only DC) on this set can be obtained by submitting prediction masks on the challenge website. The DC value drops by roughly the same amount.
The first ranked performance on the KiTS challenge leaderboard achieves 0.974 DC for kidney segmentation by employing 6-fold data augmentation and training for 1000 epochs~\cite{isensee2019attempt}. In comparison, we have achieved the reported result without any data augmentation and by training for $\frac{1}{5}^{th}$ number of epochs. This further reinforces the fast convergence of our proposed method.
 
%In comparison, training from scratch leads to a peak DC of 0.907. 

Fig.~\ref{fig:diceCurve} shows the progression of DC values over epochs for the MwS and MwoS for the validation set. For clarity of visualization, the plots are shown after a moving average operation. Relative to MwoS, MwS can be seen to yield higher DC values even in the first few epochs and converge much faster, with a higher DC value at the end of the 200 epochs.

Fig.~\ref{fig:segImage} shows the segmentation results of MwS, overlaid on the volume and ground truth. The segmentation results of MwoS overlaid on the volume, are compared on the right side. It can be seen that with MwS, the segmentation captures the fine structures as well without over-segmenting. In the case of MwoS, voxels towards the edges of the kidneys are seen to be misclassified as background while in some locations some over-segmentation can be observed.

\section{Conclusion}
We demonstrated the application of self-supervised learning for segmentation in medical imaging. A novel proxy task was used to perform self-supervised learning for segmentation. Self-supervision for segmentation is promising as it leads to faster convergence and a boost in results without using any additional data or data augmentation.
In the future, generalized criteria for the selection of a good proxy task may be formulated. The success of self-supervised learning in a more generic manner would lead towards reducing the need for large datasets for transfer learning. 
\bibliographystyle{IEEEbib}
\bibliography{main}
\end{document}